\documentclass[letterpaper, 10 pt, conference]{ieeeconf}  

\IEEEoverridecommandlockouts                              

\usepackage{amsthm}
\usepackage{amsmath,amssymb}
\usepackage[ruled,linesnumbered]{algorithm2e}
\usepackage{algorithmic}
\usepackage{mathrsfs}
\usepackage{color}
\usepackage{cite}
\usepackage{multirow}
\usepackage{url}
\usepackage{xcolor}
\usepackage{xspace}
\usepackage{graphicx}
\usepackage{subfigure} 
\usepackage{comment}

\usepackage{booktabs}
\usepackage{subcaption}

\makeatletter
\let\NAT@parse\undefined
\makeatother
\usepackage[colorlinks,
            urlcolor=blue,
            linkcolor=blue,       
            anchorcolor=blue,  
            citecolor=green        
            ]{hyperref}

\newcommand{\core}{CRPlace\xspace}

\title{\LARGE \bf
\core: Camera-Radar Fusion with BEV Representation for Place Recognition
}

\author{Shaowei Fu, Yifan Duan, Yao Li, Chengzhen Meng, Yingjie Wang, Jianmin Ji, Yanyong Zhang*
\thanks{* The corresponding author.}
\thanks{School of Computer Science and Technology, University of Science and Technology of China, Hefei, 230026, China
{\tt\small \{fushw, dyf0202, zkdly, czmeng, yingjiewang\}@mail.ustc.edu.cn, \{jianmin, yanyongz\}@ustc.edu.cn}.}%
}

\begin{document}

\maketitle
\thispagestyle{empty}
\pagestyle{empty}





\begin{abstract}

The integration of complementary characteristics from camera and radar data has emerged as an effective approach in 3D object detection. 
However, such fusion-based methods remain unexplored for place recognition, an equally important task for autonomous systems.
Given that place recognition relies on the similarity between a query scene and the corresponding candidate scene, the stationary background of a scene is expected to play a crucial role in the task. As such, current well-designed camera-radar fusion methods for 3D object detection can hardly take effect in place recognition because they mainly focus on dynamic foreground objects. 
In this paper, a background-attentive camera-radar fusion-based method, named \core, is proposed to generate background-attentive global descriptors from multi-view images and radar point clouds for accurate place recognition. To extract stationary background features effectively, we design an adaptive module that generates the background-attentive mask by utilizing the camera BEV feature and radar dynamic points. With the guidance of a background mask, we devise a bidirectional cross-attention-based spatial fusion strategy to facilitate comprehensive spatial interaction between the background information of the camera BEV feature and the radar BEV feature. As the first camera-radar fusion-based place recognition network, \core has been evaluated thoroughly on the nuScenes dataset. The results show that our algorithm outperforms a variety of baseline methods across a comprehensive set of metrics (recall@1 reaches 91.2\%).  
\end{abstract}    

\begin{figure}[htbp]
\centering
{\includegraphics[width=0.48\textwidth]{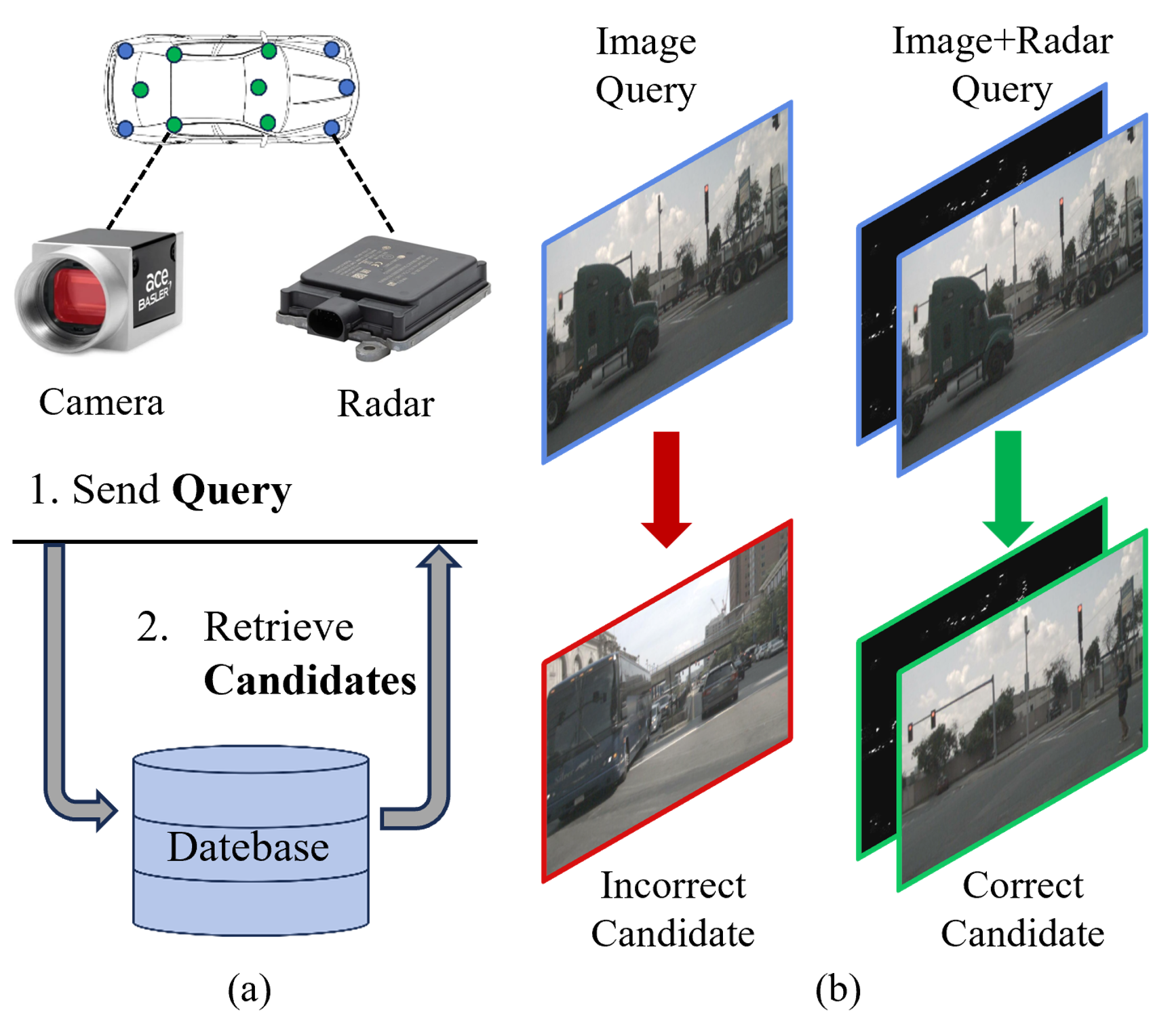}}
\caption{An illustration of (a) place recognition task with camera and radar fusion and (b) the place recognition results using image-only and fusion-based methods, respectively. Given an image query (marked in blue bounding box) that includes multiple dynamic objects, the image-based place recognition \cite{xu2023leveraging} retrieves an incorrect candidate due to the influence of dynamic objects (marked in red bounding box), while our method retrieves the correct candidate successfully (marked in green bounding box) with the image-radar query acquired from the same place. }
\label{challenge}
\vspace{-2em}
\end{figure}

\section{Introduction}

Place recognition serves as a pivotal function in autonomous driving by addressing the fundamental question of ``where am I within a predefined reference map''. As an essential component of global localization, it enables the retrieval of a place within a large map database that closely matches the current place \cite{arandjelovic2016netvlad}. Furthermore, it plays a vital role in Simultaneous Localization and Mapping (SLAM), aiding in the detection of loop closures to rectify drift and tracking errors \cite{pan2021coral}.

Cameras and LiDAR are the most commonly used sensors in place recognition. Visual place recognition \cite{arandjelovic2016netvlad, radenovic2018fine, berton2022rethinking, khaliq2022multires} has been widely studied due to the low cost and rich texture information provided by cameras. However, it is susceptible to challenges posed by visual degradation such as night, rain, and direct sunlight \cite{cai2022autoplace}. On the other hand, methods based on LiDAR \cite{uy2018pointnetvlad, chen2021overlapnet, xu2021disco, vidanapathirana2021locus, komorowski2021minkloc3d} offer improved robustness against illumination conditions but lack rich texture features. Recent efforts have sought to fuse these two modalities for place recognition \cite{lai2022adafusion, komorowski2021minkloc++, lee20232, pan2021coral, xu2023leveraging}, resulting in significant enhancements in performance and robustness. Nevertheless, these methods still fall short under adverse weather conditions such as rain, snow, and fog \cite{komorowski2021minkloc3d}. 

Unlike cameras and LiDAR, millimeter-wave radar (referred to as radar in this paper) remains nearly unaffected by harsh weather conditions and obstacles. It can provide 3D geometry information like LiDAR but is more lightweight and inexpensive. Consequently, radar is becoming an increasingly attractive sensor in autonomous driving. Due to the challenges of sparsity and noisy measurements in radar, it has become a common practice to fuse radar with camera or LiDAR to exploit their complementary characteristics for 3D object detection \cite{wang2023bi, kim2023craft, kim2023crn}. However, these methods mainly focus on pre-defined foreground objects, rendering them unsuitable for place recognition. Specifically, place recognition retrieves candidates by querying the most similar scene in the database (Fig.~\ref{challenge}(a)). In this demand, the moving objects from the scene could be misleading instead. We provide an example in Fig. \ref{challenge}(b), the image-based place recognition method~\cite{xu2023leveraging} fails to retrieve correct candidates since the disturbance from dynamic objects could not be eliminated. Therefore, \emph{it is necessary to extract global descriptors\footnote{We usually use global descriptors to describe a place.} that focus on stationary background information with the help of the velocity-sensitive radar data.}  
Although studies have attempted to exploit the unique characteristics of radar data for place recognition \cite{suaftescu2020kidnapped, barnes2020under, cai2022autoplace}, how to effectively fuse the complementary characteristics of camera and radar data while focusing on the background remains a challenge. 



In this work, we propose \core, a novel place recognition method that fuses the complementary characteristics of multi-view camera images and radar points to generate background-attentive global descriptors. Specifically, \core first engages a Background-Attentive Mask Generation (BAMG) module to adaptively create an attention mask that focuses on the stationary background feature while ignoring the dynamic feature. Guided by the background mask, a Bidirectional cross-attention-based Spatial Fusion (BSF) module is then devised to enable thorough background feature interaction and learn the soft association between camera and radar BEV features. In detail, the Radar-to-Image (R2I) fusion takes each pixel in the camera BEV feature as a query to learn spatial background information from the radar BEV feature, and the Image-to-Radar (I2R) fusion utilizes rich contextual background information from the image feature to enhance the sparse radar feature. Additionally, we establish a baseline for camera-radar fusion-based place recognition by directly combining the feature extraction module of the SOTA fusion-based detection network, i.e., BEVFusion \cite{liu2023bevfusion}, and the global descriptor aggregation module of the SOTA 360-degree visual place recognition network vDiSCO \cite{xu2023leveraging}. We refer to this baseline as BEVFusion in the remainder of the paper, which also supports camera-only and radar-only methods.

We evaluate our method on nuScenes dataset \cite{caesar2020nuscenes}. We compare \core with several state-of-the-art camera-based, radar-based, and camera-radar fusion-based methods, all of which do not take into account the influence of dynamic objects. \core outperforms these methods with significant margins (with the relative recall@1 increase of 3.6\% to 12.9\% ). We also validate the robustness of our method in rain conditions, achieving a relative recall@1 increase of 30.1\%.
In summary, our contributions are:

\begin{itemize}
\item We propose a novel and robust background-attentive Camera-Radar fusion-based place recognition method, namely \core, to combine the complementary characteristics of camera and radar in the BEV representation. To the best of our knowledge, this is the first work that effectively fuses multi-view cameras and radars for the task of place recognition.  

\item We design an adaptive background-attentive mask generation module and a bidirectional cross-attention-based spatial fusion module to learn and interact with stationary background features effectively.
\item We conduct extensive experiments on the nuScenes dataset to validate the merits of our method and show considerably improved performance. 

\end{itemize}

\begin{figure*}[htb]
\centering
{\includegraphics[width=1\textwidth]{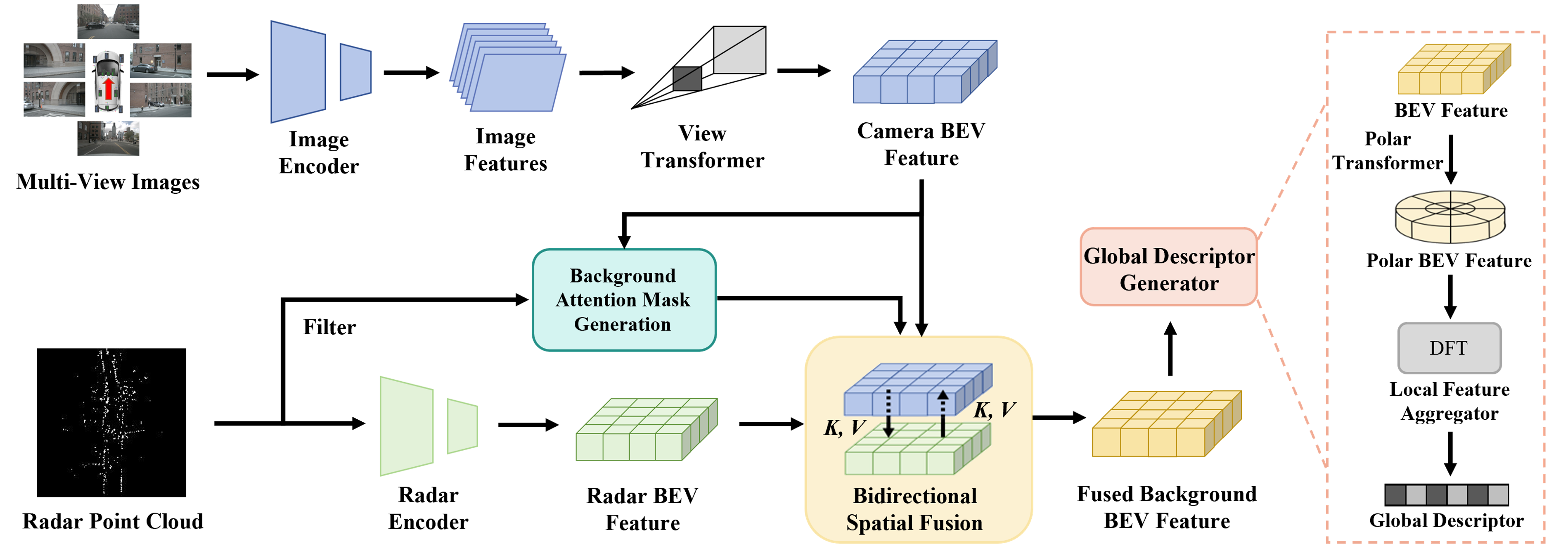}}
\caption{The network architecture of the proposed \core. Given multi-view images and radar point clouds, two modality-specific streams separately extract features and transform them into the same BEV space at first. Next, the Background-Attentive Mask Generation (BAMG) module uses radar dynamic points and camera BEV features to create a background attention mask adaptively. Then the Bidirectional Spatial Fusion (BSF) module attentively fuses background BEV features from these two modalities. Finally, the Global Descriptor Generator uses the fused BEV features to generate rotation-invariant global descriptors.}
\label{network}
\vspace{-1.5em}
\end{figure*}

\section{Related work}

\subsection{Single-modal Place Recognition}

Camera, LiDAR, and radar have all been employed for place recognition. 

\setlength{\parskip}{0.3em} \noindent\textbf{{Camera-based Methods.}} In camera-based methods, handcrafted local features \cite{valgren2007sift,bay2006surf,rublee2011orb} and their vector of locally aggregated descriptors (VLAD) \cite{arandjelovic2013all} are traditionally used for recognition, but they have been replaced by convolutional neural networks (CNNs) like VGG \cite{simonyan2014very} and AlexNet \cite{krizhevsky2012imagenet}. NetVLAD \cite{arandjelovic2016netvlad} is an end-to-end learnable method specifically designed for large-scale place recognition. It first extracts the local feature using VGG/AlexNet, followed by a differentiable VLAD layer used for local feature aggregation, which can be plugged into various learning-based feature extractors and trained through backpropagation. Similarly, Generalized-Mean Pooling (GeM) \cite{radenovic2018fine} is an efficient aggregation method, enabling the network to aggregate a compact global descriptor end-to-end. 

\setlength{\parskip}{0.3em} \noindent \textbf{{LiDAR-based Methods.}} 
Some LiDAR-based methods project point clouds to 2D structures such as Scan Context \cite{kim2018scan} and Scan Context++ \cite{kim2021scan}, while MinkLoc3D \cite{komorowski2021minkloc3d} and Locus \cite{vidanapathirana2021locus} directly operate in 3D space by discretizing it into voxel grids. Inspired by NetVLAD, PointNetVLAD \cite{uy2018pointnetvlad} combines PointNet \cite{qi2017pointnet} and NetVLAD to enable end-to-end training and extraction for the global descriptor from 3D point clouds. OverlapNet \cite{chen2021overlapnet} and DiSCO \cite{xu2021disco} try to simultaneously estimate the relative yaw between pairs of scans and their similarity. 

\setlength{\parskip}{0.3em} 
\noindent \textbf{{Radar-based Methods.}}
In radar-based methods, UnderTheradar \cite{barnes2020under} uses intermediate features as global descriptors. Kidnappedradar \cite{suaftescu2020kidnapped} exploits a variant of NetVLAD as the feature extractor to improve rotation invariance. AutoPlace \cite{cai2022autoplace} is the first work that uses single-chip automotive radar for place recognition. It first removes the dynamic points from instant Doppler measurement and then extracts spatial-temporal features from radar point clouds with a compact deep neural network. Subsequently, the obtained candidates are re-ranked using Radar Cross Section (RCS) measurement to refine the recognition accuracy.

\subsection{Multi-modal Place Recognition}

Many efforts have been made in various works towards LiDAR and camera fusion-based multi-modal place recognition. CORAL \cite{pan2021coral} first builds the elevation image from 3D points, which is then enhanced with projected RGB image features. In this way, the structural features and visual features are fused in the bird-eye view.  MinkLoc++ \cite{komorowski2021minkloc++} extracts the global descriptor for LiDAR point cloud and RGD image separately and fuses them in the last channel. Considering the fact that the importance of camera and LiDAR varies as the environment changes, AdaFusion \cite{lai2022adafusion} tries to learn the adaptive weights for both image and point cloud features. However, methods based on camera-radar fusion for place recognition have not yet received any attention.


\subsection{Rotation Invariance}

Rotation invariance is crucial in place recognition. Bird's Eye View (BEV) representation has a natural advantage in achieving rotation invariance, which has been widely used in LiDAR-based methods \cite{kim2021scan, xu2021disco, chen2021overlapnet}. OverlapNetVLAD \cite{fu2023overlapnetvlad} is a coarse-to-fine place recognition framework that efficiently uses BEV features to perform loop closure. Previous visual place recognition methods use single-view image to extract the global descriptor, which fail to retrieve the correct candidates when revisiting the same place from different perspectives. Pioneering works LSS \cite{philion2020lift}, BEVFusion \cite{liu2023bevfusion} and BEVFormer \cite{li2022bevformer} have demonstrated that aggregating features from multi-view images into a unified BEV representation can significantly improve detection and segmentation performance. Inspired by these methods, a recent work, called vDiSCO \cite{xu2023leveraging}, proposes a method to employ BEV representation for 360-degree visual place recognition, achieving remarkable results. vDiSCO extracts BEV features from multi-view images based on BEVFormer, then combines the polar transformation and the Discrete Fourier transform to aggregate rotation-invariant global descriptors from BEV features. It also supports the vision-LiDAR fusion method. This method has effectively demonstrated the advantages of BEV representation in the task of place recognition.


\section{Method}

In this work, we present \core, a background-attentive camera-radar fusion network for place recognition. As shown in Fig. \ref{network}, multi-view images and radar point clouds are separately fed into the camera feature stream and radar feature stream to extract their BEV features. Next, we involve a Background-Attentive Mask Generation (BAMG) module to create a background attention mask adaptively by combining camera BEV features and radar dynamic points. Then a Bidirectional cross-attention-based Spatial Fusion (BSF) module is devised with the guidance of the attention mask to interact with the background features attentively. Subsequently, the rotation-invariant global descriptor is generated according to the method in vDiSCO \cite{xu2023leveraging}. Following this way, a place can be represented by a background-attentive global descriptor, denoted as $D_p$, so that we can generate a global descriptor for each place in a given map to build a database $\{D_i\}$. We also generate the global descriptor for the current query place, say $D_q$. By comparing the Euclidean distance between the query place descriptor and the place descriptor in the database, the current place can be finally recognized as the one in the map with the minimal difference, denote the $i^*$th place as: 
\begin{equation}
      i^* = \arg \mathop {\min }\limits_i \left\| {{D_q} - {D_i}} \right\|_2.
\end{equation}

Below we will present the details of \core. 


\subsection{Modality-Specific BEV Feature Encoding}
\subsubsection{Multi-View Image Feature Encoding}
Following BEVFusion \cite{liu2023bevfusion}, we use Swin-T \cite{liu2021swin} as the image backbone to encode the multi-view images into deep features. The Feature Pyramid Network (FPN) \cite{lin2017feature} is then applied to fuse the multi-scale features. An adaptive average pooling layer is applied to better align these features. To transform these image features from 2D coordinate into 3D ego-car coordinate, we apply the view transformer proposed in \cite{philion2020lift} to explicitly predict the depth distribution of each pixel. Then, the image-view features can be projected onto the predefined point cloud and a pseudo voxel $V \ {\in} \ R^{X{\times}Y{\times}Z{\times}C}$ can be generated according to camera extrinsic parameters and the predicted image depth. Note that the depth prediction module may produce inaccurate depth, leading to the projection of image features to incorrect BEV positions. Therefore, we use ground-truth depth derived from LiDAR point clouds to supervise the depth distribution prediction during training, following BEVDepth \cite{li2023bevdepth}. Subsequently, a BEV pooling is applied to reshape the pseudo voxel into a BEV feature map $F_C \ {\in} \ R^{C{\times}H{\times}W}$. 

\subsubsection{Radar Feature Encoding}
We preprocess the radar point cloud into a feature set containing the 2D coordinate $(x,y)$, radial velocity $(v_x,v_y)$, radar cross-section $rcs$, dynamic property $dynProp$, cluster validity state $invalid \ state$ and timestamp $t$. Five frames of radar scans within the same sample are stacked into a full-view radar point cloud in the LiDAR coordinate. To avoid overly sparse inputs, we further accumulate a sequence of radar point clouds into one frame. The dynamic points are identified and removed from the radar point cloud based on radial velocity, as described in \cite{cai2022autoplace}. Then, we exploit a PillarFeatureNet \cite{lang2019pointpillars} to extract radar features which directly converts the radar input to a pseudo image in the BEV space. This encoder considerably alleviates the computation of sparse radar data to traverse the BEV plane in the absence of vertical information. Naturally, the radar BEV features $F_R\ {\in} \ R ^ {C{\times}H{\times}W}$ are generated after a linear transformation, where $C$, $H$, and $W$ are equal to $F_C$.

\begin{figure}[!t]
\centering
{\includegraphics[width=0.48\textwidth]{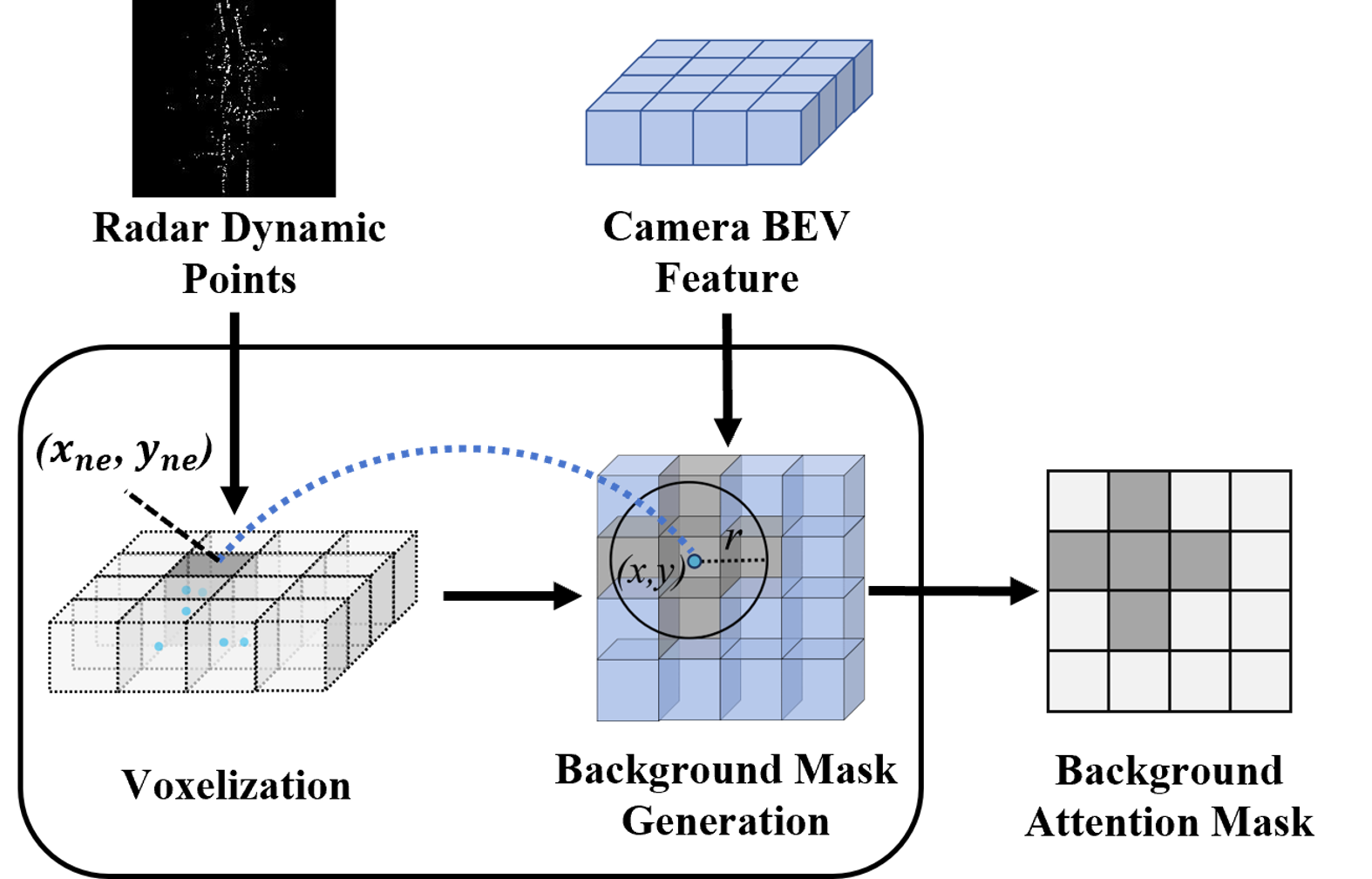}}
\caption{An illustration of the BAMG module. All dynamic points are selected from radar point clouds and voxelized into a grid. Then the radar voxel grid and camera BEV feature are utilized to generate the background attention mask adaptively according to their positional relationships. $(x_{ne},y_{ne})$ denotes the non-empty voxel.}
\label{bamg}
\vspace{-1.5em}
\end{figure}

\begin{figure*}[htbp]
\centering
{\includegraphics[width=1\textwidth]{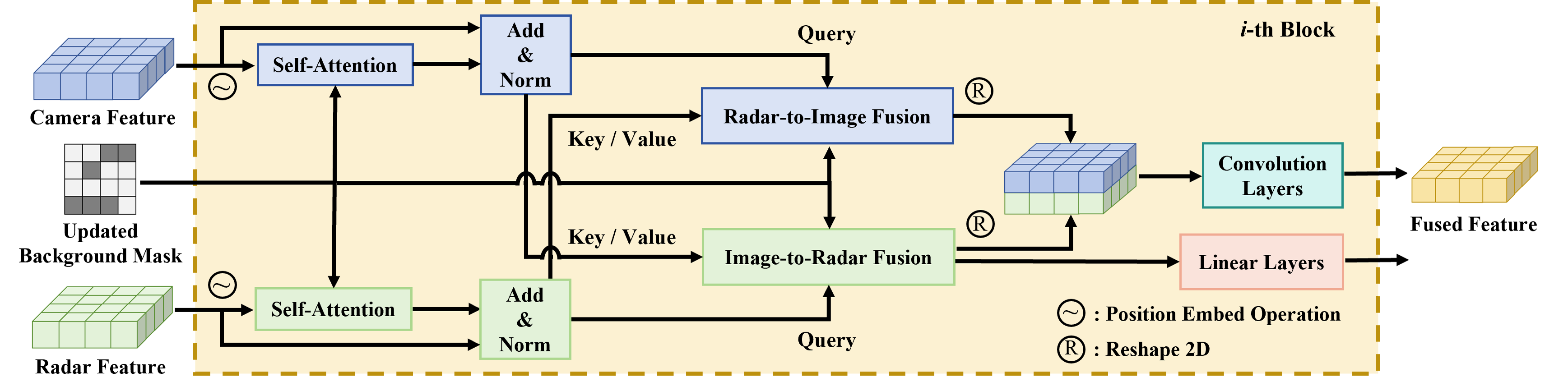}}
\caption{An illustration of the Bidirectional Spatial Fusion block. Take camera BEV feature, radar BEV feature, and background attention mask as input, a Self-Attention module is first applied to these two features respectively. Then a Radar-to-Image Fusion and an Image-to-Radar Fusion are used for bidirectional spatial interaction. Finally, a convolution-based fusion operation is performed. A linear layer is used to generate the input of radar feature for the next block.}
\label{fuser}
\vspace{-1.5em}
\end{figure*}

\subsection{Background-Attentive Mask Generation}

Before feature fusion, it is necessary to distinguish which areas in the BEV features represent dynamic objects and which represent stationary backgrounds. In our BAMG module, a background attention mask $M \ {\in} \ \{ 0, 1 \} ^ {H \times W}$ of the same size as the camera BEV feature is defined to prevent attention for dynamic features, where $H = W = 128$. As shown in Fig. \ref{bamg}, we first retain all dynamic points in the radar point cloud and then perform voxelization to generate a voxel grid $V \ {\in} \ R ^{H \times W} $. For each non-empty grid $V(x_{ne}, y_{ne})$, it means that there is a dynamic object near the position $(x_{ne},y_{ne})$. The corresponding position in the camera BEV feature is also considered to have a dynamic object because the voxel grid and the camera BEV feature have been transformed into the same BEV coordinate system. Furthermore, the incorporation of explicit depth supervision during the view transformer process of the camera feature stream further enhances the spatial alignment of the camera BEV feature and radar voxel grid. While we have a general idea of the approximate position of the dynamic object, its spatial extent is not clear. In other words, it is uncertain which pixels belong to the dynamic features. As a result, we set a learnable parameter $r$ to allow the network to adaptively learn the areas occupied by dynamic features. Specifically, for each non-empty voxel grid $V(x_{ne}, y_{ne})$, if the Euclidean distance between the coordinate $(x,y)$ of camera BEV feature pixel $F_C(x, y)$ and $(x_{ne}, y_{ne})$ is less than the threshold $r$, it is considered that this feature pixel belongs to the dynamic feature, and the corresponding position of background attention mask $M(x, y)$ will be set to 0. This can be represented using the following formula:

\begin{equation}
M(x, y) =
\begin{cases}
0,  & \text{if \ $\left\| {(x, y) - (x_{ne}, y_{ne})} \right\|_2 \leq r$ } \\
1, & \text{else}
\end{cases},
\end{equation}
where $x,x_{ne} \ {\in} \ H, \ y,y_{ne}\ {\in} \ W$ and $r$ is initialized to 0.5.


\subsection{Bidirectional Spatial Fusion}

While it is possible to identify stationary background features using the generated background mask, simply fusing camera features and radar features through element-wise addition or concatenation would result in spatial misalignment and significant feature wastage due to the sparsity of radar measurements. For this reason, our BSF module aims to fully interact the complementary characteristics and learn the soft-association between camera features and radar features attentively under the guidance of the background attention mask. The BSF module is composed of a stack of 3 identical blocks. As illustrated in Fig. \ref{fuser}, a block consists of four parts: a Self-Attention module, a Radar-to-Image Fusion module, an Image-to-Radar Fusion module, and a convolution-based fusion operation. 

\subsubsection{Radar-to-Image Fusion}
Radar-to-Image (R2I) Fusion provides the spatial background information of radar features to image features. The positional embedding operation is applied based on their corresponding BEV spatial coordinates before fusion. To enhance the intrinsic representation capability of camera background features, given a $C$ dimensional camera BEV feature map $F_C \ {\in} \ R^{C{\times}H{\times}W}$ as queries $Q^I$, we first perform the deformable self-attention (DSA) for each query $Q^I_p \ {\in} \ Q^I$ as follow:
\begin{equation}
    DSA(Q^I_p) = DefAttn(Q^I_p, p, V^I),
\end{equation}
where $Q^I_p$ represents the camera BEV query at point $p = (x_p,y_p)$, and $V^I \ {\in} \ Q^I$ is the features with background attention mask $M(x, y) = 1$. 
Next the radar BEV feature $F_R'\ {\in} \ F_R$ with $M(x, y) = 1$ is used as key and value to perform the Radar-to-Image deformable cross-attention as follow:
\begin{equation}
    R2I(Q^I_p, F_R') = \sum\limits_{V \in F_R'} DefAttn(Q^I_p, p, F_R').
\end{equation}

\subsubsection{Image-to-Radar Fusion}
Image-to-Radar (I2R) Fusion module utilizes rich contextual background information from image features to complement sparse radar features. Similarly, a radar BEV feature $F_R \ {\in} \ R^{C{\times}H{\times}W}$ is used as query $Q^R_p$ to perform deformable self-attention as follow:
\begin{equation}
    DSA(Q^R_p) = DefAttn(Q^R_p, p, V^R),
\end{equation}
where $Q^R_p$ represents the radar BEV query at point $p = (x_p,y_p)$, and $V^R \ {\in} \ Q^R$ is the features with $M(x, y) = 1$. 
Then the Image-to-Radar deformable cross-attention is performed as follow:
\begin{equation}
    I2R(Q^R_p, F_C') = \sum\limits_{V \in F_C'} DefAttn(Q^R_p, p, F_C'),
\end{equation}
where $F_C' \ {\in} \ F_C$ is the camera BEV feature with $M(i, j) = 1$.
Subsequently, the output of I2R is fed into a linear layer for the next $(i+1)$-th block.

\subsubsection{Convolution-based Fusion}
Designed for 2D structures, convolution kernels are better at extracting local spatial correlations than 1D attention. Therefore, the outputs of R2I fusion and I2R fusion are transformed to image style again and concatenated along the channel dimension, then sent to the convolution block, expressed as follows:
\begin{equation}
    F^i_{out} = H(R2I(Q^I, F_R') \ \oplus I2R(Q^R, F_C')),
\end{equation}
where $F^i_{out}$ is the output of the $i$-th block, and $H$ represents the convolution-based fusion operation. In this way, multiple blocks increase the fitness of $F_C$ and $F_R$, and the background BEV features can be enhanced gradually.  

\subsection{Global Descriptor Generator}
Following vDiSCO \cite{xu2023leveraging}, the fused background BEV features are fed into a Global Descriptor Generator to aggregate rotation-invariant global descriptors. Firstly, the polar transformation is applied to transform BEV features into the polar coordinate system, and then DFT is performed on the polar BEV features to achieve rotation invariance. Specifically, the rotation invariance is realized by the translation invariant property of the magnitude spectrum on polar BEV, where the translation indicates the rotation in the original BEV.

\section{Experiments}

\subsection{NuScenes Dataset}
NuScenes \cite{caesar2020nuscenes} is the first dataset for large-scale environments with multi-modal sensors, including LiDAR, camera, and radar. There are six camera sensors installed in front, front left, front right, back, back left, and back right parts of the vehicle, and five radar sensors installed at the front, left, right, and back, covering a 360° FOV. Following AutoPlace \cite{cai2022autoplace}, we use the largest split, \textit{Boston} split to train and evaluate our \core. This split is divided into \textit{database set}, \textit{training query set}, \textit{validation query set}, and \textit{test query set} containing 6312, 7075, 924, and 3696 multi-view images and radar point clouds, respectively. See AutoPlace \cite{cai2022autoplace} for more details about the dataset.

\subsection{Implementation Details}

For multi-view images, we set the image size to $256 \times 704$, and the recognition range of the BEV grid is $(-51.2, 51.2)m$ for the $X, Y$ axis, and $(-10, 10)m$ for $Z$ axis. To densify the radar point cloud, we follow typical data pre-processing of \cite{lin2020depth} to concatenate the nearest six radar point clouds using ground truth ego-motion. The radar voxel size is set to $(0.8, 0.8, 8)m$. In order to convert camera features and radar features into a unified BEV space, we transfer the 2D position and velocity of radar points from the radar coordinates to the LiDAR coordinates. 

For the network training, we follow the common practice \cite{xu2023leveraging, cai2022autoplace} to adopt metric learning with triplet margin loss. Multi-view images and a corresponding radar point cloud form a mini-batch. Each batch consists of several mini-batches that can be divided into a query, positive and negative samples. Following the scale of AutoPlace \cite{cai2022autoplace}, we
regard places in the database that are within the radius=9 m area
to the query as positive samples, while those are outside the
radius=18 m area as negative samples. The loss term is given as:
\begin{equation} \small
{\rm{L}} = \sum\limits_k {\max \{ \left\| {f(q),f(p)} \right\|_2 - \left\| {f(q),f({n^k})} \right\|_2 + m,0\} },
\end{equation}
where $f(\cdot)$ denotes the network mapping a mini-batch to a
feature vector, $\left\| \cdot \right\|_2 $ means Euclidean distance, $q$ is the query sample, $p$ is the best positive matching sample, $n^k$ is the negative sample, $m = 0.1$ is the predefined margin, and $k = 10$ is the number of negative samples. We use a batch size of 4 and SGD with an initial learning rate of 0.001, momentum of 0.9, and weight decay of 0.001.  We decay the learning rate by 0.5 every 5 epochs. 

\subsection{Evaluation Metrics}
We use \textit{Recall@N} \cite{arandjelovic2016netvlad}, precision-recall curve \cite{hou2018evaluation}, $max \ F_1$ \cite{suaftescu2020kidnapped} and average precision (AP) \cite{hou2018evaluation} to evaluate the performance of different place recognition methods. \textit{Recall@N} measures the percentage of successfully localized queries using the top $N$ candidates retrieved from the database. Localization is successful if one of the top $N$ retrieved candidates is within $d$ meters of the ground truth. In our experiments, $d$ is set to $9m$.


\subsection{Comparison with State-of-The-Art Methods}

We compare our method with both single-modality (camera or radar) and multi-modality (camera-radar) place recognition methods, including:

\begin{itemize}
\item Camera-based methods: NetVLAD \cite{arandjelovic2016netvlad}, vDiSCO \cite{xu2023leveraging}, and BEVFusion \cite{liu2023bevfusion} (camera-only). The input of NetVLAD is the front-view images, and the others are multi-view images.
\item Radar-based methods: AutoPlace \cite{cai2022autoplace}, and BEVFusion \cite{liu2023bevfusion} (radar-only). 
\item Camera-Radar fusion-based method: BEVFusion \cite{liu2023bevfusion}.
\end{itemize}

We adapt the implementation of the above works to the settings of the nuScenes dataset. 

\begin{figure}[!t]
\centering
{\includegraphics[width=0.5\textwidth]{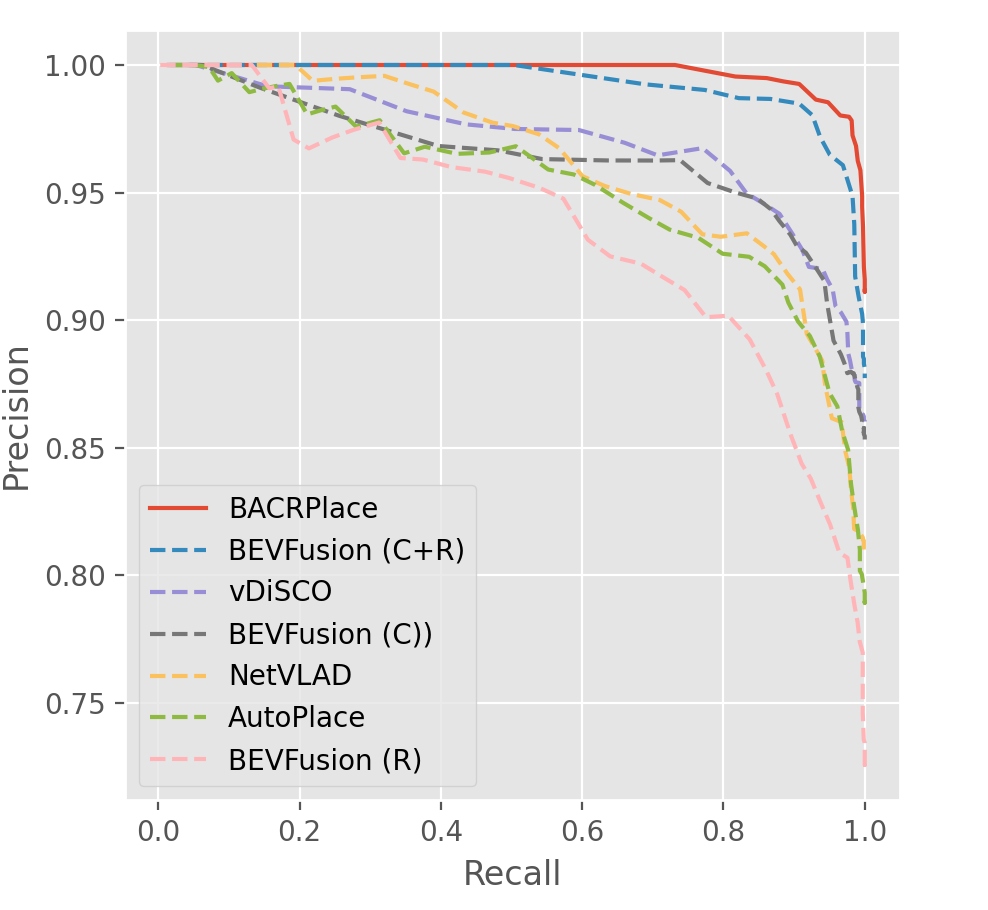}}
\caption{Precision-recall curve of SOTA methods on the nuScenes dataset.}
\label{pr_curve}
\vspace{-1.5em}
\end{figure}

Table \ref{tab:results} shows the comparison results of place recognition methods on the nuScenes dataset. For visual place recognition, the multi-view methods with BEV representation vDiSCO and BEVFusion outperform the single-view method NetVLAD, achieving 86.0\% and 85.5\% recall@1, respectively. This is because BEV representation can provide rotation-invariant global descriptors. For radar-based place recognition, they both underperform compared to visual methods because images have stronger representational capabilities. AutoPlace is the SOTA radar-based method as it converts radar point clouds into images. Instead, BEVFusion (R) has inferior performance because it directly extracts features from radar point clouds. Notably, BEVFusion (C+R) outperforms those based on a single modality in all metrics, which indicates the effectiveness of the fusion approach. Our method further improves recall@1, $max \ F_1$, and AP to 91.2\%, 0.96, and 0.98, respectively. We can also observe a similar trend from Fig. \ref{pr_curve} that camera-radar fusion-based place recognition methods are outperformed by single modality-based methods. Still, \core exceeds the others by a significant margin  (from 3.6\% to 12.9\% relative increase of recall@1).

\begin{figure}
    \centering
    \subfigure[Query]{
        \centering
        \includegraphics[width=0.28\linewidth]{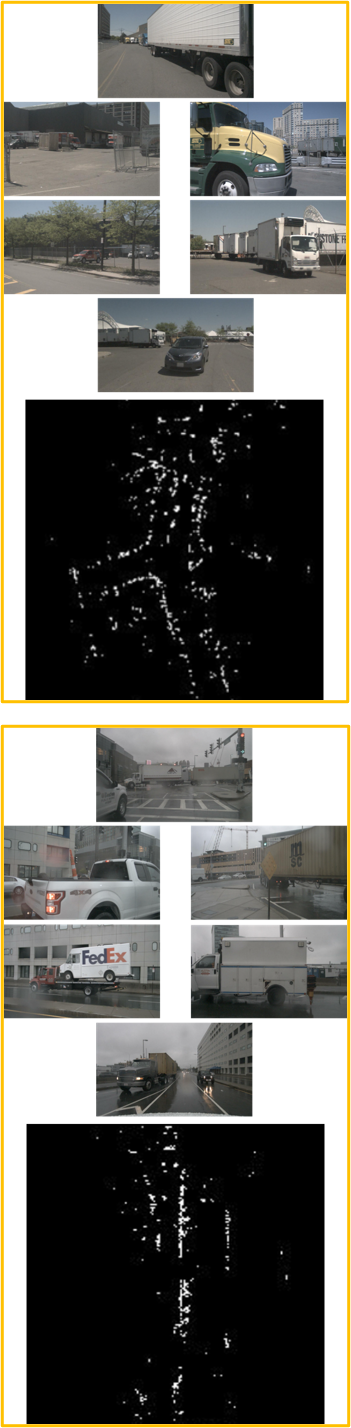}
            \label{fig:a}
    }
    \subfigure[BEVFusion]{
        \centering
        \includegraphics[width=0.28\linewidth]{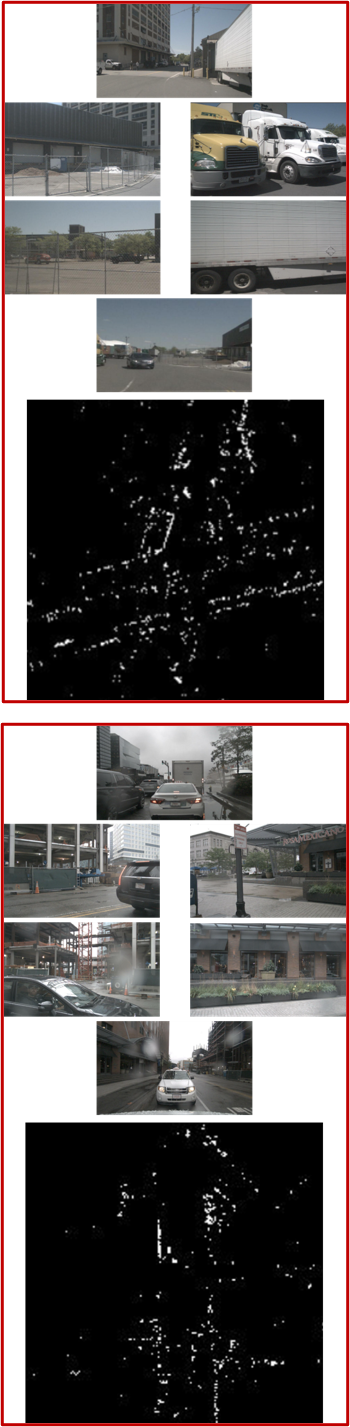}
        \label{fig:b}
    }
    \subfigure[\core]{
        \centering
        \includegraphics[width=0.28\linewidth]{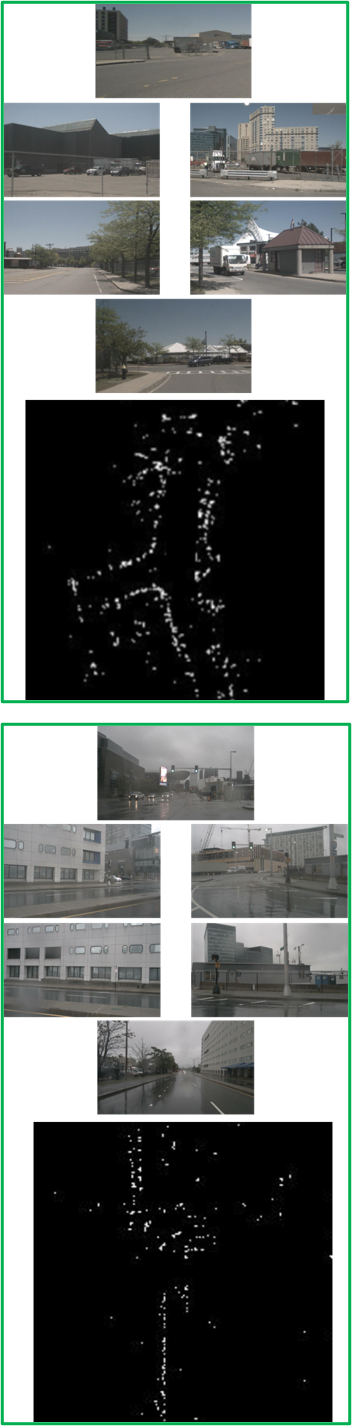}
        \label{fig:c}
    }
    \caption{Qualitative comparison between BEVFusion and our \core. (a) are queries influenced by (1) dynamic objects (first row) and (2) dynamic objects + rain conditions (second row). (b) is an incorrect retrieval using BEVFusion, and (c) is a correct retrieval using \core.}
    \label{qualitative}
    \vspace{-1.5em}
\end{figure}

We also provide qualitative analysis in Fig. \ref{qualitative}. As we can see, when a query is surrounded by many dynamic objects (first row) and also in the rain conditions (second row), BEVFusion will retrieve a false positive, while \core can still retrieve the correct match.

\begin{table}
    \centering
    \caption{Comparison results of place recognition methods on the nuScenes dataset. C denotes camera, and R denotes radar.}
    \begin{tabular}{@{}lccccc@{}}
        \toprule
        Method & Modality & Recall@1/5/10 & $max \ F_1$ & AP\\
        \midrule
        NetVLAD \cite{arandjelovic2016netvlad} & C & 80.8/86.2/87.6 & 0.91 & 0.95\\
        vDiSCO \cite{xu2023leveraging} & C & 86.0/88.6/89.2 & 0.95 & 0.96\\
        BEVFusion \cite{liu2023bevfusion} & C & 85.5/87.9/88.7 & 0.94 & 0.96\\
        \midrule
        AutoPlace \cite{cai2022autoplace} & R & 77.8/82.3/83.7 & 0.94 & 0.97\\
        BEVFusion \cite{liu2023bevfusion} & R & 72.5/79.0/80.8 & 0.89 & 0.92\\
        \midrule
        BEVFusion \cite{liu2023bevfusion} & C+R & 88.0/89.9/90.6 & 0.95 & 0.96\\
        \core (ours) & C+R & \textbf{91.2}/\textbf{92.6}/\textbf{93.3} & \textbf{0.96} & \textbf{0.98} \\
        \bottomrule
    \end{tabular}
    \label{tab:results}
\end{table}


\subsection{Ablation Study}

To understand how each module in \core affects the place recognition performance, we conduct ablation studies by evaluating different groups shown in Table \ref{tab:ablation}.

\textit{Method (a)} is our camera-radar fusion-based baseline BEVFusion, which achieves recall@1 of 88.0\%, recall@5 of 89.9\%, and recall@10 of 90.6\%.

\textit{Method (b)} extends \textit{(a)} by simply adding the BAMG module. The background mask is directly added to the camera and radar BEV features, and then the original feature fusion module in BEVFusion \cite{liu2023bevfusion} is performed. However, this method does not yield a significant improvement in performance. We believe this is because the original feature fusion module does not effectively leverage the background mask to learn background features.

\textit{Method (c)} extends \textit{(a)} by replacing the original feature fusion module with our BSF module. Even without explicit guidance from a background mask, this method still improves recall@1 by 2.4\%. This shows that our BSF module has powerful feature fusion capabilities.

\textit{Method (d)} is our \core. By combining the BAMG module and the BSF module, it achieves a gain of 3.2\% for recall@1 compared to BEVFusion. This indicates that our BSF module can effectively utilize the background mask to fuse background features attentively.

We also investigate the impact of different feature aggregation methods on \core, including NetVLAD \cite{arandjelovic2016netvlad}, GeM \cite{radenovic2018fine}, and DFT \cite{xu2023leveraging}, which are used for generating global descriptors. As shown in Table \ref{tab:pool}, unsurprisingly, DFT outperforms other methods as it generates rotation-invariant global descriptors.

\begin{table}
    \centering
    \caption{Ablation Study of \core.}
    \resizebox{\linewidth}{!}{
    \begin{tabular}{@{}cccccc@{}}
        \toprule
        Method & BAMG & BSF & Recall@1/5/10 & $max \ F_1$ & AP\\
        \midrule
        (a) &  &  & 88.0/89.9/90.6 & 0.95 & 0.96 \\
        (b) & \checkmark &  & 88.2/89.9/90.5 & 0.94 & 0.96 \\
        (c) & & \checkmark & 90.4/91.3/92.2 & 0.96 & 0.97\\
        (d) & \checkmark & \checkmark & \textbf{91.2}/\textbf{92.6}/\textbf{93.3} & \textbf{0.96} & \textbf{0.98} \\

        \bottomrule
    \end{tabular}
    }
    \label{tab:ablation}
\end{table}


\begin{table}
    \centering
    \caption{Comparative study of feature aggregation methods on \core.}
    \begin{tabular}{lcccc}
    \toprule
        Aggregation & Recall@1 & $max \ F_1$ & AP\\
    \midrule
        NetVLAD \cite{arandjelovic2016netvlad} & 87.4 & 0.95 & 0.96\\
        GeM \cite{radenovic2018fine} & 89.7 & 0.96 & 0.96 \\
        DFT \cite{xu2023leveraging} & \textbf{91.2} & \textbf{0.96} & \textbf{0.98} \\
    \bottomrule
    \end{tabular}
    \label{tab:pool}
    \vspace{-1em}
\end{table}


\begin{table}[htbp]
    \centering
    \caption{Comparison results of place recognition methods in rain conditions.}
    \resizebox{\linewidth}{!}{
    \begin{tabular}{@{}lccccc@{}}
        \toprule
        Method & Modality & Recall@1/5/10 & $max \ F_1$ & AP\\
        \midrule
        NetVLAD \cite{arandjelovic2016netvlad} & C & 65.8/75.4/77.1 & 0.86 & 0.93 \\
        vDiSCO \cite{xu2023leveraging} & C & 70.2/76.6/80.4 & 0.89 & 0.94\\
        BEVFusion \cite{liu2023bevfusion} & C & 69.9/76.4/80.1 & 0.88 & 0.94\\
        \midrule
        AutoPlace \cite{cai2022autoplace} & R & 75.7/82.0/83.5 & 0.94 & 0.96 \\
        BEVFusion \cite{liu2023bevfusion} & R & 71.6/77.4/79.7 & 0.88 & 0.93\\
        \midrule
        BEVFusion \cite{liu2023bevfusion} & C+R & 83.8/87.3/87.9 & 0.93 & 0.96\\
        \core (ours) & C+R & \textbf{85.6}/\textbf{88.6}/\textbf{89.1} & \textbf{0.95} & \textbf{0.96} \\
        \bottomrule
    \end{tabular}
    }
    \label{tab:rain}
    \vspace{-1em}
\end{table}

\subsection{Comparison in Rain Conditions}

To verify the robustness of our method under varying environmental conditions, we filter the rain-affected samples from the \textit{Boston} split in nuScenes dataset as the validation set for additional evaluation. As indicated in Table \ref{tab:rain}, visual place recognition methods demonstrate considerable performance deterioration in these conditions. For instance, the recall@1 of NetVLAD drops to only 65\%, while vDiSCO and BEVFusion (C) show similar diminished effectiveness, each achieving around a 70\% recall@1 rate. In contrast, radar-based methods and camera-radar fusion-based methods maintain relatively strong performance. Notably, the efficacy of radar-based approaches has now surpassed that of purely visual methods in rainy conditions. Our methodology stands out in this challenging environment, showcasing an improvement in recall@1 ranging between 2.1\% to 30.1\% compared to other methods. This demonstrates its considerable potential for reliable performance in rain-impacted scenarios.

\section{Conclusion}

In this paper, we introduce \core, a background-attentive bidirectional fusion method that fuses the complementary camera and radar data for improving place recognition. Unlike existing camera-radar fusion schemes that focus on dynamic features in 3D object detection, we leverage the dynamic properties of radar points to adaptively discern which features belong to the stationary background. Subsequently, a bidirectional cross-attention mechanism is employed to interactively fuse background features from both the camera and radar. With our background-attentive bidirectional fusion method, \core outperforms earlier schemes for place recognition on nuScenes dataset.

\bibliographystyle{IEEEtran}
\bibliography{IEEEabrv, references}

\end{document}